\documentclass[runningheads]{llncs}

\usepackage[mobile]{accv}

\usepackage{accvabbrv}

\usepackage{graphicx}
\usepackage{booktabs}

\usepackage[accsupp]{axessibility}  % Improves PDF readability for those with disabilities.

\usepackage{pifont} 
\newcommand{\xmark}{\ding{55}}

\usepackage[pagebackref,breaklinks,colorlinks,citecolor=accvblue]{hyperref}

\usepackage{xcolor}

\usepackage{booktabs}    % 美观表格
\usepackage{colortbl}    % 允许表格颜色
\usepackage{xcolor}      % 颜色支持
\usepackage{afterpage}
\definecolor{LightBrown}{RGB}{222,184,135} % 浅棕色

\usepackage{orcidlink}
\usepackage[misc]{ifsym}

\begin{document}

% \title{Localized Feature Discrimination Graph Convolution Network for Skeleton-based Action Recognition} 
\title{Synchronized and Fine-Grained Head for Skeleton-Based Ambiguous Action Recognition}


\author{Hao Huang$^\star$\inst{1}\orcidlink{0009-0001-9791-4452} \and
Yujie Lin$^\star$\inst{1} \and
Siyu Chen$^\star$\inst{1} \and
Haiyang Liu\textsuperscript{\dag}\inst{1} }

\footnotetext[1]{Equal contributions}
\footnotetext[4]{Corresponding author. Email: \href{mailto:haiyangliu@bjtu.edu.cn}{haiyangliu@bjtu.edu.cn}}

\authorrunning{H.~Huang et al.}

\institute{School of Computer and Information Technology\\Beijing Jiaotong University, Beijing 100044, China\\
}

\maketitle

\begin{abstract}

Skeleton-based action recognition using Graph Convolutional Networks (GCNs) has achieved remarkable performance, but recognizing ambiguous actions, such as "waving" and "saluting", remains a significant challenge. Existing methods typically rely on a serial combination of GCNs and Temporal Convolutional Networks (TCNs), where spatial and temporal features are extracted independently, leading to an unbalanced spatial-temporal information, which hinders accurate action recognition. Moreover, existing methods for ambiguous actions often overemphasize local details, resulting in the loss of crucial global context, which further complicates the task of differentiating ambiguous actions. To address these challenges, we propose a lightweight plug-and-play module called Synchronized and Fine-grained Head (SF-Head), inserted between GCN and TCN layers. SF-Head first conducts Synchronized Spatial-Temporal Extraction (SSTE) with a Feature Redundancy Loss (F-RL), ensuring a balanced interaction between the two types of features. It then performs Adaptive Cross-dimensional Feature Aggregation (AC-FA), with a Feature Consistency Loss (F-CL), which aligns the aggregated feature with their original spatial-temporal feature. This aggregation step effectively combines both global context and local details, enhancing the model’s ability to classify ambiguous actions. Experimental results on NTU RGB+D 60, NTU RGB+D 120, NW-UCLA and PKU-MMD I datasets demonstrate significant improvements in distinguishing ambiguous actions. Our code will be made available at \href{https://github.com/HaoHuang2003/SFHead}{https://github.com/HaoHuang2003/SFHead}.

\end{abstract}

\section{Introduction}
\label{sec:intro}

Human action recognition is a critical technology that aims to understand human movements by analyzing video or sensor data, and it finds wide applications in human-computer interaction, intelligent surveillance, and related domains. Skeleton-based action recognition, a specialized subset of this field, focuses on using skeletal representations of human bodies for recognition tasks.

Recent years have seen a transition from handcrafted features \cite{choutas2018potion, zhouyan2019pa3d} to deep learning approaches involving Recurrent Neural Networks (RNNs) \cite{ke2017new, song2017end, liu2017skeleton, li2018independently} and Convolutional Neural Networks (CNNs) \cite{hou2016skeleton, li2017joint, ke2017skeletonnet}. However, these early methods primarily emphasized either temporal sequence modeling or simplistic joint relationship extraction, failing to fully capture the intricate dependencies between joints, which are crucial for recognizing complex human actions.

Yan et al. \cite{yan2018spatial} were first to introduce GCN into skeleton action recognition and propose the spatial temporal graph convolutional network (ST-GCN), using the non-Euclidean space to store the skeleton information. Nevertheless, ST-GCN employed a fixed topology across all layers, which failed to fully exploit vital information such as bone length and orientation, restricting its adaptability to complex motions. To address this limitation, subsequent research \cite{shi2019two,chen2021channel} proposed adaptive and dynamic topologies, enabling models to adjust the correlations between joints. But they are still lack of ability to capture the details between distant joints. Lee et al. \cite{Lee_2023_ICCV} proposed a Hierarchically Decomposed Graph Convolutional Network (HD-GCN) that efficiently decomposes joints into hierarchical subgroups, allowing better representation of structurally distant joints. However, existing approaches typically combine GCNs and TCNs in a serial manner, where spatial and temporal features are extracted independently. This often results in an imbalance, as the interplay between spatial and temporal features is not fully exploited, limiting the model's accuracy.

While most existing methods effectively capture global information, they often struggle with complex or ambiguous skeleton actions due to their relatively weaker ability to preserve fine-grained local details. Ambiguous actions, such as "waving" versus "saluting" or "reading" versus "writing," often exhibit similar motion trajectories with subtle differences. Distinguishing these actions requires maintaining a balance between fine-grained local details and global context, which current methods fail to achieve effectively. 
Zhou et al. \cite{zhou2023learning} first proposed the discriminative Feature Refinement Head (FR-Head), which employs contrastive learning to enhance ambiguous action recognition. This was later extended by \cite{chen2024dstc, Huang2024}. However, these methods primarily focused on refining local details, often at the expense of global context. Moreover, they introduced additional parameters, increasing computational complexity.

To overcome such limitations, we propose the Synchronized and Fine-grained Head (SF-Head), a lightweight plug-and-play module designed to improve the discriminative capacity of skeleton-based ambiguous action recognition models. SF-Head consists of two key components: Synchronized Spatial-Temporal Extraction (SSTE) and Adaptive Cross-dimensional Feature Aggregation (AC-FA). First, SSTE module extracts spatial and temporal feature in a synchronized manner to preserve important relationships between them. Moreover, we propose a Feature Redundancy Loss (F-RL) to balance these two types of features. Then, the spatial-temporal feature with original feature are then fed into AC-FA module which is adaptive aggregation across multiple dimensions—channel, spatial and temporal, combining both global context and local details. To further improve the aggregation, we introduce a Feature Consistency Loss (F-CL), which aligns the aggregated features with their original extracted spatial-temporal feature, serving as a supervisory signal to prevent the loss of important feature information during aggregation.

Our SF-Head has fewer than 0.01M parameters and can be integrated into any GCN-based architecture, providing a lightweight yet effective solution for improving the feature discriminative power for ambiguous actions. Importantly, SF-Head is only used during training, with no additional computational cost during inference.

Our major contributions are summarized as follows:
\begin{itemize}
\item We propose SF-Head, a lightweight, plug-and-play module with few than 0.01M parameters. It can be embedded into any GCN-based network without additional computational cost during inference. 
\item SF-Head improves skeleton-based action recognition of ambiguous actions by synchronizing spatial-temporal feature extraction with F-RL to balance spatial and temporal features, followed by adaptive cross-dimensional aggregation with F-CL to align cross-dimensional features, effectively combining global context with local details. 
\item Extensive experiments on four benchmark datasets (NTU RGB+D 60, NTU RGB+D 120, NW-UCLA and PKU-MMD I) demonstrate that our method significantly improves classification accuracy in distinguishing ambiguous actions.
\end{itemize}

\section{Related Work}

\subsection{Skeleton-based Action Recognition}

Skeleton-based action recognition has gained significant traction due to its robust anti-jamming properties and lower data complexity compared to RGB-based approaches \cite{peng2016bag, li2020directional, zhang2019two}. Early methods typically relied on handcrafted features, such as joint distances, angles, and speeds, and utilized RNNs \cite{su2020predict} or CNNs \cite{choutas2018potion, zhouyan2019pa3d} to process these features. In RNN-based methods, skeletal data is modeled as joint-coordinate vectors over time, whereas CNN-based approaches represent these coordinates as pseudo-images for feature extraction. The simplicity of CNN training has made it more popular compared to RNNs.

Yan et al. \cite{yan2018spatial} propose ST-GCN using Graph Convolution Networks (GCNs), which first makes the conventional skeleton action recognition handled in a non-Euclidean space, effectively capturing the complex spatial relationships between different body parts. However, a major limitation of ST-GCN was the use of a fixed topology across all network layers, which constrained the model's ability to adaptively capture vital skeletal information, such as bone length and joint orientations.

To address the limitations of a fixed topology, subsequent works like Shi et al. \cite{shi2019two} and Li et al. \cite{li2019actional} proposed adaptive graph structures that learn joint associations dynamically, enabling the network to capture complex spatial interdependencies. However, they may still overlook potential relations between distant joints, reducing the network's ability to represent the global structure effectively, especially for actions involving multiple body parts.

To enrich the representation capabilities, Chen et al. \cite{chen2021channel} introduced a topology refinement module to improve the feature extraction process, while Chi et al. \cite{chi2022infogcn} proposed a multimodal skeletal representation to enhance the spatial awareness of each joint. More recently, Lee et al. \cite{Lee_2023_ICCV} introduced a Hierarchically Decomposed Graph Convolutional Network (HD-GCN) that hierarchically decomposes skeletal joints to better represent structurally distant joints, allowing improved inter-joint connectivity and information flow. Zhou et al. \cite{zhou2024blockgcn} further refined topological representation by incorporating bone connectivity encoding and persistent homology, aimed at addressing potential topology degradation during feature extraction. Qu et al. \cite{qu2024llms} proposes using LLMs for skeleton-based action recognition, leveraging a linguistic projection process to treat skeleton sequences. Abdelfattah et al. \cite{abdelfattah2024maskclr} focuses on improving the robustness of transformer-based models by introducing a masked contrastive learning strategy. Additionally, Myung et al. \cite{myung2024degcn} proposed the Deformable Graph Convolutional Network (DeGCN), which adaptively captures the most informative joints by learning deformable sampling locations.

While these works have significantly advanced the modeling of skeletal data, most existing methods use a serial combination of GCN and TCN to extract spatio-temporal features, potentially leading to unbalanced feature extraction. In this work, we propose SSTE module in SF-Head, that synchronizes the spatio-temporal feature extraction with F-RL to preserve the balanced relationship between two types of features.

\subsection{Ambiguous Action Recognition}

Ambiguous action recognition has gained traction due to its importance in distinguishing visually similar but contextually different human activities, such as "writing" and "reading." Early attempts in recognizing such ambiguous samples are rooted in fine-grained image classification, where minor visual differences between classes are key to distinguishing them \cite{Veeriah_2015_ICCV, Dubey_2018_ECCV, Zhuang_Wang_Qiao_2020}. Zhou et al. \cite{zhou2023learning} were the first to explore ambiguous action classification for skeleton-based data. They proposed the FR-Head, which used contrastive learning to dynamically refine fuzzy action samples within the feature space, improving the ability to differentiate visually similar movements. However, FR-Head may introduce substantial additional parameters during training and suffers from loss of global context when over-focusing on local differences.To address such limitations, Chen et al. \cite{chen2024dstc} designed a differential spatiotemporal correlation model that captures both local joint connections and non-adjacent temporal frames, focusing on subtle differences in highly similar actions. Meanwhile, Huang et al. \cite{Huang2024} introduced the Ambiguous Exclusion and Contrast Learning (AEC) module to exclude uncertain samples from training, prioritizing confident samples to stabilize the feature distribution. However, exist approaches overemphasize
local details, resulting in the loss of essential global context, which can significantly hinder ambiguous action recognition. 

In this paper, AC-FA module are proposed in SF-Head to combine global context with local details with F-CL to align the aggregated features with their original decoupled spatiotemporal feature, improving the model's ability to differentiate ambiguous actions.

\begin{figure*}[!t]
    \centering
    \includegraphics[scale=0.19]{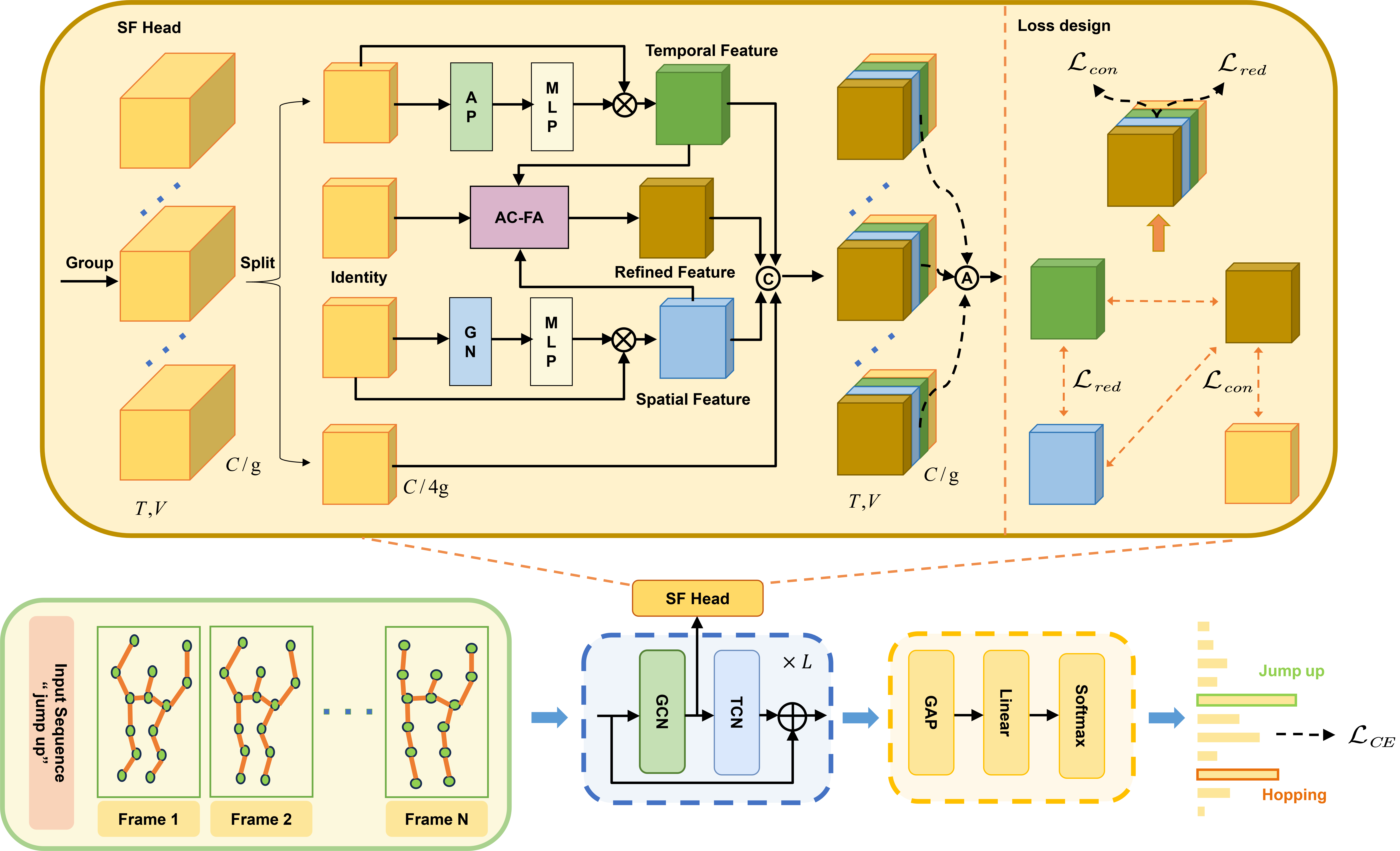}
    \caption{\textbf{The overall framework of our proposed method.}
We propose a lightweight, plug-and-play module SF-Head inserted between GCN and TCN, designed to synchronizing spatial-temporal decoupling with $\mathcal{L}_{red}$ to balance spatial and temporal features, followed by performing adaptive cross-dimensional feature aggregation(AC-FA) module with $\mathcal{L}_{con}$ to align the refined feature with channel, temporal and spatial feature, combining global context and local details.}
    \label{SF-Head}
\end{figure*}

\section{Methodology}
The overview of our method is depicted in \cref{SF-Head}. The backbone of our model, described in \cref{sec3.1}, consists of $L$ basic blocks constructed using a combination of GCNs and TCNs. In \cref{sec3.2}, we propose a lightweight, plug-and-play module SF-Head which contains SSTE and AC-FA module inserted between GCNs and TCNs. The loss design, as shown in \cref{SF-Head}, will be introduced in \cref{sec3.3}.

\subsection{Backbone}
\label{sec3.1}
The input to our model is a sequence of 3D skeletal data represented as \( X \in \mathbb{R}^{3 \times T \times V} \), where \( T \) and \( V \) denote the number of frames and the number of joint nodes, respectively. We propose the SF-Head module, which is embedded in the model's backbone consisting of 10 basic blocks based on \cite{chen2021channel}, constructed using a combination of GCNs and TCNs. Specifically, GCNs capture spatial features by applying learnable topological graphs over the spatial domain, as described in \cref{sec4.5.3}, while TCNs extract temporal features using 1D convolutions over the temporal domain.

After the feature map passes through all 10 basic blocks, a global average pooling layer is applied to compress the feature map into a 1D high-level feature vector. This vector is then passed through a fully connected (FC) layer, and the final softmax activation outputs a probability distribution for action classification.

It is important to note that our method is not dependent on the specific implementation of the backbone. The basic block can be substituted with any GCN-based architecture, such as those in \cite{Lee_2023_ICCV, zhou2024blockgcn, chi2022infogcn}, as demonstrated in \cref{sec4.4}.

\subsection{Synchronized and Fine-grained Head(SF-Head)}
\label{sec3.2}
Our primary goal is to enhance the performance of skeleton-based models in recognizing ambiguous actions that exhibit subtle differences and are prone to misclassification. To achieve this, we propose a lightweight, plug-and-play module termed Synchronized and Fine-Grained Refinement Head (SF-Head), designed to synchronize the spatial and temporal extraction and aggregate cross-dimensional features within the backbone. 

% Specifically, SF-Head first decouples the feature map into spatial and channel features in a synchronized manner with a Feature Redundancy Loss (F-RL) to balance these two components. Then, we propose the adaptive cross-dimensional feature aggregation(AC-FA) embedded within SF-Head to aggregate multiple dimensional features from channel, spatial and temporal with a Feature Consistency Loss (F-CL)  to align the aggregated features with their original decoupled components.

Importantly, SF-Head is only used during training, introducing no additional computational costs during inference. 
\subsubsection{Synchronized Spatial-Temporal Extraction(SSTE)}

In skeleton-based action recognition, the traditional sequential approach of employing GCNs for spatial feature extraction followed by TCNs for temporal modeling results in a fundamental limitation: asynchronous feature extraction. By treating spatial and temporal features independently in a serial manner, these models generate unaligned and incomplete feature representations. Such disjointed extraction impairs the model's capacity to fully capture the intricate spatio-temporal relationships required for distinguishing subtle variations in complex actions, leading to critical information loss.

To address this issue, we propose a synchronized Spatial-Temporal extraction(SSTE) module that mitigates the information loss associated with traditional asynchronous pipelines to improve the discriminative capability of action recognition.

As shown in \cref{SF-Head}, the input of SF-Head is the raw feature map  $X\in \mathbb{R}^{C\times T\times V}$, which is firstly divided into $g$ groups along the channel dimension, i.e., $X=\left[ X_1,...,X_g \right] ,X_k\in \mathbb{R} ^{C/g\times T\times V}$ and then each grouped feature $X_k$ is split into four parallel branches along the channel dimension, i.e., $X_{k1,}X_{k2,}X_{k3,}X_{k4,}\in \mathbb{R} ^{C/4g\times T\times V}$, which is inspired by  \cite{Lee_2023_ICCV}. In this section, we mainly focus on the $X_{k1}$ and $X_{k3}$, termed $X_t$ and $X_s$ repectively, which are fed into two parallel branches. The synchronized feature representations are the temporal feature $f_t$ and the spatial feature $f_s$. 

\begin{equation}
\begin{aligned}
f_t &= \sigma ( \phi_t(GP(X_{t})) ) \cdot X_t, \\
f_s &= \sigma ( \phi_s(GN(X_{s}) ) \cdot X_s.
\end{aligned}
\end{equation}

where $AP$ is averaging pooling along the temporal dimension and $GN$ is Group Norm  \cite{wu2018group}. $\phi_t$ and $\phi_s$ are temporal and spatial feature head for feature extraction, respectively. $\sigma$ is the sigmoid activation. \newline

\subsubsection{Adaptive Cross-Dimensional Feature Aggregation(AC-FA)}

While the SSTE module within SF-Head effectively extract temporal and spatial representations, it falls short to balance the global context and local details. 

For instance, actions like "waving" and "saluting" exhibit highly similar spatial configurations, yet their differences lie in subtle joint movements and temporal progressions. SSTE module tend to capture broad spatial-temporal patterns, often treating these minor differences as noise, which severely limits the model's ability to discriminate between such ambiguous actions. Therefore, build on \cite{misra2021rotate}, we introduce the Adaptive Cross-Dimensional Feature Aggregation(AC-FA) model embedded in SF-Head, aimed at aggregating the synchronized spatial-temporal feature representations with the original feature to combine the global context and local details. 

\begin{figure*}[ht!]
    \centering
    \includegraphics[scale=0.25]{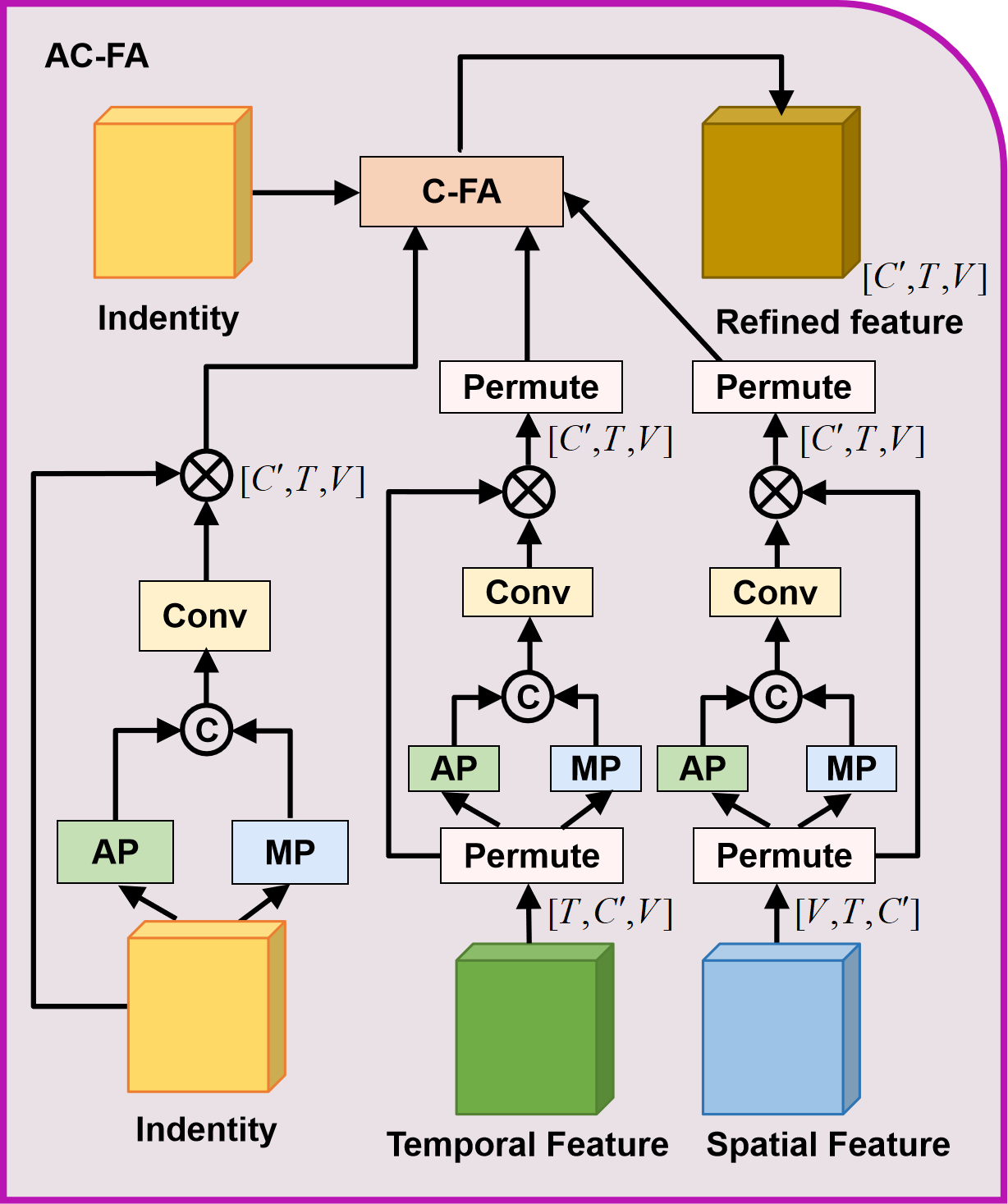}
    \caption{Adaptive Cross-Dimensional Feature Aggregation(AC-FA) module}
    \label{AC-FA}
\end{figure*}

As \cref{AC-FA} shows, AC-FA aggregates cross-dimensional features by extracting key elements from the identity feature $X_{k2}$ termed $f_c$, temporal feature $f_t$ and spatial feature $f_s$.

\textbf{Adaptive Channel Dimension Aggregation(ACDA)}

The original feature $f_c\in \mathbb{R}^{C'\times T \times V}$ retains the raw channel information which is critical for capturing the diverse activation responses across different joints and frames. In the context of ambiguous action recognition, such as distinguishing between similar motions like 'waving' and 'saluting,' the subtle differences often lie in the specific patterns of activation across different body parts, especially the upper body. By aggregating channel-level features, the module can selectively emphasize the most relevant joints or body parts, which are critical for distinguishing these fine-grained variations in action.

To fully leverage channel information, AC-FA module applies average pooling $(AP)$ and max pooling $(MP)$ across the channel dimension to capture both general and highly salient features:
\begin{equation}
f_{c}^{*}=concat\left( AP\left( f_c \right) ,MP\left( f_c \right) \right)
\end{equation}
where $concat(\cdot)$ represents the concatenate operation; $f_{c}^{*}$ is enriched channel feature map which is subsequently refined through 2D convolution, followed by batch normalization and sigmoid activation to produce channel attention weight $\omega_c$:
\begin{equation}
\omega_c=\sigma \left( \psi _c\left( f_{c}^{*} \right) \right)
\end{equation}
where $\psi_c$ is the 2D convolution operation. The adaptive channel weight $\omega_c \in \mathbb{R}^{1 \times T \times V}$ is applied to $f_c$, allowing SF-Head to emphasize channels that are most relevant for discriminating between actions, ensuring that the most discriminative channel-level information is preserved.

\textbf{Adaptive Temporal Dimension Aggregation(ATDA)}

Capturing the temporal dynamics is crucial, especially for actions that involve similar postures but different timing. To effectively capture temporal variations, we reshape the temporal feature map $f_t \in \mathbb{R}^{C' \times T \times V}$ into $\hat{f_t} \in \mathbb{R}^{T \times C' \times V}$, emphasizing the temporal variation of each channel and joint across frames. Then, we extract both the average and maximum trends across the temporal dimension to analyze temporal relationships:
\begin{equation}
\hat{f_{t}^{*}}=concat\left( AP\left( \hat{f_t} \right) ,MP\left( \hat{f_t} \right) \right)
\end{equation}
where $\hat{f_t^*}$ represents the enriched temporal representation. It is then passed through a 2D convolutional layer, followed by batch normalization and a sigmoid activation to generate the frame-level attention weight $\omega_t \in \mathbb{R}^{1 \times T \times V}$. 
\begin{equation}
\omega_t=\sigma \left( \psi _t\left( \hat{f_{t}^{*}} \right) \right)
\end{equation}
where $\psi_t$ is the 2D convolution operation; $\omega_t$ is then applied to the reshaped temporal feature map $\hat{f_t}$, adaptively emphasizing frames that carry critical temporal cues. This process ensures that temporal dependencies and important frame-level variations are preserved to capture subtle timing differences that may distinguish ambiguous actions.

\textbf{Adaptive Spatial Dimension Aggregation(ASDA)}

The spatial relationships among joints are critical for distinguishing actions that share similar global postures. For joint-level aggregation, the spatial feature $f_s \in \mathbb{R}^{C' \times T \times V}$ is reshaped into $\hat{f_s} \in \mathbb{R}^{V \times T \times C'}$, which focuses on variations across joints for each frame. $\hat{f_s}$ is passed through $AP$ and $MP$ layer, aiming to capture both general joint behavior as well as the most significant activations:
\begin{equation}
\hat{f_{s}^{*}}=concat\left( AP\left( \hat{f_s} \right) ,MP\left( \hat{f_s} \right) \right)
\end{equation}
where $\hat{f_{s}^{*}}$ is the enriched joint-level feature followed by a 2D convolutional layer, batch normalization, and sigmoid activation, generating joint-level attention weights $\omega_v$:
\begin{equation}
\omega_s=\sigma \left( \psi _s\left( \hat{f_{s}^{*}} \right) \right)
\end{equation}
Here, $\psi_s$ represents the 2D convolution operation. $\omega_s$ is then applied to the reshaped spatial feature $\hat{f_s}$, allowing AC-FA module to adaptively emphasize joints that contribute most to distinguishing the action.

\textbf{Cross-Dimensional Feature Aggregation(C-FA)}

Finally, the AC-FA module aggregates the refined features from all three dimensions $f_c,f_t$ and $f_s$ together with the original feature $X_{k4}$, denoted as $f_o$, to produce a comprehensive aggregated feature representation $f_a$:
\begin{equation}
f_a = {\sum_{i \in \{c, t, s, o\}} \eta_i \left( \omega_i f_i + \mathbb{I}_{\{t,s\}}(i) \left(\overline{\omega_i\hat{f}_i} - \omega_i f_i \right) \right)}
% f_a={ \left(\eta_c f_c\omega _c+ 
%   \eta_t\overline{\hat{f_{t}}{\omega}_{t}} + \eta_s\overline{\hat{f_{s}}{\omega}_{s}}+\eta_of_o \right)}
\label{eq.8} 
\end{equation}
where $\eta_i$ is hype-parameter to control the adaptive contribution of each feature map to the final aggregated feature representation. For $ i \in \{t, s\}$, the indicator function $ \mathbb{I}_{\{t,s\}}(i) $ activates and replaces the original product $ \omega_i f_i $ with its transformed counterpart $\overline{\omega_i\hat{f}_i} $. Furthermore, to simplify \eqref{eq.8} we would obtain $f_a$:
\begin{equation}
% f_a={\eta \left( y_c + 
%   \overline {y_t} + \overline {y_s} + y_o\right)}
f_a = \sum_{i \in \{c, t, s, o\}} \eta_i \left( y_i + \mathbb{I}_{\{t,s\}}(i) \left( \overline{y_i} - y_i \right) \right)
\label{MVFA}
\end{equation}
where $y_i$ represent the feature map after the adaptive cross-dimension aggregation process; $\overline{y_t}$ and $\overline{y_s}$ in \eqref{MVFA} denote the reshaped versions of $y_t$ and $y_s$ to match the original input shape $(C',T,V)$.

The Adaptive Cross-dimensional Feature Aggregation (AC-FA) module, as shown in \cref{AC-FA}, adaptively integrates the most discriminative aspects from all three dimensions—channel, temporal, and spatial—while effectively combining global context and local details. This leads to a more nuanced and discriminative understanding of complex and ambiguous actions.

\subsection{Loss Design for Feature Learning}
\label{sec3.3}

\subsubsection{Feature Redundancy Loss(F-RL)}

When the spatial and temporal feature representations extracted by the SSTE module become overly correlated or too independent, the distribution of information may become imbalanced, thus hindering the model's ability to effectively capture the subtle yet crucial distinctions between similar actions. To address this issue, building on the work \cite{wang2022renovate}, we propose the Feature Redundancy Loss (F-RL), which aims to balance the spatial and temporal feature representations. Meanwhile, We adopt a modified cosine distance $d$ to quantify the alignment between these two types of feature maps :
\begin{equation}
d\left( u,v \right) =1-e^{-\alpha \left( \frac{uv^T}{\left\| u \right\| _2\left\| v \right\| _2}+1 \right)}
\end{equation}
where $u,v$ represent 1D vectors which are flattened feature maps; $\alpha$ is hype-parameter to regulate the smoothness of the distance function and $\left\| \cdot \right\|$ is $L_2$ norm.

The proposed F-RL is defined as:
\begin{equation}
\mathcal{L} _{red}=-\frac{1}{2N}\sum_{i=1}^N{{\log \left( \frac{e^{m\cdot d(f_s^i,f_t^i)/\tau}}{\sum_{j=1}^N{e^{m\cdot d(f_s^i,f_t^i)/\tau}}} \right)}}
\label{F-DL}
\end{equation}

where $N$ is the number of samples in a batch; $f_t^i$ and $f_s^i$ represent the temporal and spatial feature of sample i, respectively; $\tau$ is the temperature hyper-parameter that controls the sharpness of the distribution; $m$ is a scaling factor that adjusts the contrast.

By balancing the spatial and temporal features, F-RL ensures that both feature types complement each other without one becoming overly dominant. This allows the model to focus on the most relevant temporal and spatial cues, which is especially useful for distinguishing between actions that have similar global postures but different timing or joint activations, such as "reading" versus "writing."

\subsubsection{Feature Consistency Loss}

To address the challenge of retaining both local and global feature properties during the aggregation process, we introduce the Feature Consistency Loss (F-CL). The primary goal is to ensure that refined features maintain sufficient alignment with their original versions, thereby preventing over-specialization that may lead to a loss of critical context. Ambiguous actions such as "waving" and "saluting" require not only capturing nuanced details but also preserving global consistency to enable effective differentiation.

Inspired by the soft-margin mechanism \cite{liang2017soft} commonly employed to maintain optimal boundaries in classification tasks, we aim to ensure that the final cross-dimensional aggregated feature $f_a$ does not significantly deviate from its original version $F={[f_c,f_s,f_t,f_o]}$. 

We encourage the aggregated feature $f_a$ to remain close to their original feature set $F$ by assigning a soft-margin penalty. To further illustrate the idea, We construct a feature consistency loss(F-CL) term as follows:
\begin{equation}
\mathcal{L} _{con}=\frac{1}{N}\sum_{i=1}^N{\sum_F{\left( \log \left( e^m+(1-e^m)d(f_{a}^{i},F^i) \right) \right)}}
\end{equation}
where $f_a^i$ and $F^i$ represent the final aggregated and original feature of sample i, respectively; $m$ is a margin parameter controlling the penalty imposed on deviations from the original feature. The term $\mathcal{L}_{con} \in [0,m]$ because of $d\in[0,1]$ to effectively limits drastic shifts while allowing the SF-Head to aggregate important details.

However, solely minimizing this loss could lead to undesirable edge cases: features that are either excessively similar to their original state, implying under-refinement, or significantly divergent, resulting in a loss of global consistency.

Thus, to prevent both extreme similarity, which suggests excessive retention of original features and excessive deviation indicating potential loss of global context, a compensation term $\phi^i$ is introduced:
\begin{equation}
\phi^i = 
\left\{ 
\begin{array}{ll}
(1-d(f_{a}^{i},F^i))^{\gamma}\log(d(f_{a}^{i},F^i)), & \text{if } \min(d(f_{a}^{i}, F^i), 1-d(f_{a}^{i}, F^i)) > \epsilon ; \\ 
\hfill 0, & \text{otherwise.}
\end{array} 
\right.
\end{equation}
where $\epsilon$ is a small positive threshold value to filter out trivial deviations; $\gamma$ is a hyper-parameter to encourage adaptive data selection for uncertain samples.  

By integrating the additional penalty term $\phi^i$ into F-CL, we ensure that the refined features do not either overemphasize the original representation, or drifting excessively from it, thereby enhancing both global feature integrity and consistency across three dimensions. For example, when the modified cosine distance $d(f_{a}^{i},F^i)$ converges to 0 or 1, $\phi^i$ reaches the minimum value 0 to avoid overcorrection and maintain a balance between feature consistency and diversity. This approach is particularly effective for ambiguous actions such as "writing" and "reading" where the model must balance global context and local detail.

Finally, the F-CL can be rewritten as 
\begin{equation}
\mathcal{L} _{con}=\frac{1}{N}\sum_{i=1}^N{\sum_F{\left( \log \left( e^m+(1-e^m)d(f_{a}^{i},F^i) \right) + \phi^i \right)}}
\label{F-CL}
\end{equation}

\subsection{Training Objective}

Once the final feature map has been processed through the final FC layer, it is transformed into a logits vector $h\in \mathbb{R}^Z$, where $Z$ represents the number of action classes. ${p}_{i}^{c}$ is the probability score of sample $i$ belonging to class $c$, which can be computed by a softmax function:
\begin{equation}
{p}_{i}^{c}=\small{\frac{e^{h_{i}^{c}}}{\sum_{l=0}^{Z-1}{e^{h_{l}^{c}}}}}
\end{equation}

where $h_{i}^c$ denotes the $c-$th element of the logits vector $h$ for sample $i$. Then, the Cross-Entropy (CE) loss can be calculated:
\begin{equation}
\mathcal{L} _{\mathrm{CE}}=-\frac{1}{N}\sum_{i=1}^N{\sum_{c=1}^Z{y_{i}^{c}\log\mathrm{(}p_{i}^{c})}}
\end{equation}
where $y_i^c \in \mathbb{R}^Z$ represents the one-hot vector indicating the ground truth of action class of sample $i$. And $y_{i}^c=1$ if $c$ is the true class label for sample $i$.

Finally, we combine the CE loss with our proposed F-CL and F-RL to create the full learning objective:
\begin{equation}
\mathcal{L} _{\mathrm{total}}=\mathcal{L} _{\mathrm{CE}}+\lambda _{con}\cdot \mathcal{L} _{con}+\lambda _{red}\cdot \mathcal{L} _{red}
\end{equation}
where $\mathcal{L}_{con}$ and $\mathcal{L}_{red}$ are defined in \eqref{F-CL} and \eqref{F-DL}. $\lambda_{con}$ and $\lambda_{red}$ are hyper-parameters used to balance the contributions of F-CL and F-RL.

\section{Experiments}
\label{sec:blind}

\subsection{Dataset}

\subsubsection{NTU RGB+D 60.} NTU RGB+D 60
\cite{shahroudy2016ntu}, comprising 4 million frames and 56880 video sequences, is a large-scale dataset for human action recognition. There are 60 distinct action classes in the dataset such as daily, mutual, and health-related actions, which was gathered from 40 different subjects. The authors established two standards: (1) cross-subject (X-sub): there are 40 distinct subjects. Half of them is used for training group and rest for test group. (2) cross-view (X-view): there are three different camera views. Cameras 2 and 3 is chose for training, and camera 1 for testing.

\subsubsection{NTU RGB+D 120.} An expanded version of NTU RGB+D, NTU RGB+D 120 \cite{liu2019ntu}, has 8 million frames and over 114 thousand video samples. The dataset collected from 106 distinct subjects is divided into 120 different action classes. The writers established two standards: (1) cross-subject (X-sub): there are 106 distinct subjects. Half of them is used for training group and rest for test group. (2) cross-setup (X-set): samples with even setup IDs is chose for training, and samples with odd setup IDs for testing.
Samples for training are selected with even setup IDs, whereas samples with odd setup IDs are selected for testing.

\subsubsection{NW-UCLA.} The Northwestern-UCLA dataset \cite{wang2014cross} comprises 1494 video sequences spanning 10 action categories, with each action performed by 10 distinct subjects. The dataset is recorded simultaneously from 3 different viewpoints using Kinect cameras. Following the standard evaluation protocol, the first two of the three camera views are utilized for training, while the remaining camera view serves as the testing set.

\subsubsection{PKU-MMD I.}
The PKU-MMD I dataset \cite{liu2017pku} consists of 1,076 long video sequences covering 51 action classes. These actions were performed by 66 subjects from three different camera viewpoints using a Kinect v2 sensor. Following a cross-subject evaluation protocol, the 66 subjects are divided into training and testing sets, consisting of 57 and 9 subjects, respectively.

\subsection{Implementation Details} 

Our experiments employ CTR-GCN \cite{chen2021channel} as the backbone and are conducted on NVIDIA RTX A6000 with PyTorch \cite{paszke2017automatic} deep learning framework. Stochastic Gradient Descent (SGD) \cite{ruder2016overview} is applied as the optimizer in the NTU-RGB+D and PKU-MMD I datasets while the momentum is set to 0.9, along with a weight decay of 0.0004. We set the batch size to 64 and total number of epochs to 90, utilizing a warm-up strategy to avoid initial instability in the first 5 epochs. Cosine annealing is adopted as the decay strategy, parameterizing the initial and terminal learning rate as 0.1 and 0.0001. For the NW-UCLA dataset, we set the batch size to 16. The hyper-parameters in the methodology are adjusted to: $m = 0.4$, $\gamma = 2$, $\lambda_{con} = 0.2$, $\lambda_{red} = 0.1$.

\subsection{Ablation Study}

On NTU-RGB+D 120’s X-sub benchmark, extensive experiment results are presented as a typical verification example to highlight the effectiveness of various sub-modules in the proposed SF-head approach. As presented in \cref{Ablation1}, it determines the optimal parameters for loss function. As shown in \cref{Ablation2} and \cref{Ablation3}, we randomly aggregate one or two proposed sub-modules each time into the backbone to quantify their significant degrees respectively.

\subsubsection{Hyper-parameter Settings of Loss Functions (F-RL and F-CL)}
The results of the hyper-parameter settings of the system are shown in \cref{Ablation1}. First, three different values (1, 0.1, 0.01) are tried for $\lambda_{con}$ and $\lambda_{red}$ under the condition that $m = 0.5$, $\gamma = 2$. It can be seen that the performance will be poor if the value of $\lambda_{con}$ or $\lambda_{red}$' is too high or too low. Next, $\lambda_{con}$ is fine-tuned, determining $\lambda_{con} = 0.2$ and $\lambda_{red} = 0.1$ to be the more appropriate values. Then, $m$ and $\gamma$ are assigned higher or lower values around 0.5 and 2 respectively (No.7~No.9), demonstrating that the accuracy enhances for moderate values. After that, $m$ is fine-tuned to 0.4, making the accuracy peak. Therefore, the best accuracy is achieved with $m = 0.4$, $\gamma = 2$, $\lambda_{con} = 0.2$, $\lambda_{red} = 0.1$, saving as the configuration for the following experiments.

Additionally, the relatively small impact of hyper-parameter adjustments on the performance further highlights the robustness and stability of the system, demonstrating the ability of our approach to generalize well without being overly sensitive to hyper-parameter choices. This feature significantly reduces the need for extensive hyper-parameter tuning, thus improving the efficiency and reliability of the system in real-world applications.

\begin{table*}[!t]
\centering
\caption{Results of hyper-parameter experiments in F-RCL on NTU-RGB+D 120 dataset's X-sub benchmark, using only joint modality as input. F-RCL represents the combination of F-RL and F-CL.}
\label{Ablation1}
\setlength{\tabcolsep}{12pt}
\scalebox{0.8}{
\begin{tabular}{ccccccc}
\toprule
{\textbf{No.}} & \textbf{$m$} & \textbf{$\gamma$} & \textbf{$\lambda_{con}$} & \textbf{$\lambda_{red}$} &\textbf{Acc (\%)} \\
\midrule
1   &  0.5  &  2  &  1  &  1  &  85.48 \\
2   &  0.5  &  2  &  0.1  &  0.1  &  85.80 \\
3   &  0.5  &  2  &  0.01  &  0.01  &  85.34 \\
4   &  0.5  &  2  &  0.5  &  0.1  &  85.52 \\
5   &  0.5  &  2  &  0.2  &  0.1  &  85.96 \\
6   &  0.5  &  2  &  0.05  &  0.1  &  85.47 \\
7   &  1.0  &  3  &  0.2  &  0.1  &  85.76 \\
8   &  0.6  &  2  &  0.2  &  0.1  &  85.92 \\
9   &  0.2  &  1  &  0.2  &  0.1  &  85.64 \\
10  &  0.8  &  2  &  0.2  &  0.1  &  85.88 \\
11  &  \textbf{0.4}  &  \textbf{2}  &  \textbf{0.2}  &  \textbf{0.1}  &  \textbf{86.04} \\
12  &  0.2  &  2  &  0.2  &  0.1  &  85.93 \\
\bottomrule
\end{tabular}
}
\end{table*}

\subsubsection{Effectiveness of Sub-modules in SF-head}
The results of ablation studies between modules in SF-head are displayed in \cref{Ablation2}. We divided SF-Head into 3 sub-modules: SSTE, AC-FA, F-RCL. They are responsible for synchronized spatial-temporal feature extraction, feature aggregation and feature learning with key discriminative features preserved respectively. The base backbone CTR-GCN is formed by removing all three sub-modules. Compared with the base backbone (No.1), it can be seen from No.2 to No.4 that each sub-module has a positive contribution to improving the accuracy, with AC-FA and F-RCL exerting relatively significant effects. Furthermore, when it comes to adding two sub-modules from the base backbone (No.6 to No.8), all results outperform those adding a single sub-module. Among them, the combination of SSTE and AC-FA, along with SSTE and F-RCL, shows remarkable synergistic effect, exceeding the accuracy gain brought about by individual sub-modules. This demonstrates that SSTE extends the feature perspective and provides more potential feature inputs for both AC-FA's aggregation processing and F-RCL's relationship evaluation, performing a comprehensive and detailed feature representation. However, the combination of AC-FA and F-RCL shows a slight redundancy effect because of their identical focus on alignment of channel, spatial, and temporal dimensions to ensure the preservation of original features during the aggregation process. In addition, when adopting all three sub-modules, the accuracy achieves a significant improvement of 1.32\%.

\begin{table*}[!t]
\centering
\caption{Ablation results of sub-modules in SF-head on NTU-RGB+D 120 dataset's X-sub benchmark, using only joint modality as input. F-RCL represents the combination of F-RL and F-CL.}
\label{Ablation2}
\scalebox{0.9}{
\begin{tabular}{ccccc} % 修改为 ccccc
\toprule
\textbf{No.} & \textbf{SSTE} & \textbf{AC-FA} & \textbf{F-RCL} & \textbf{Acc (\%)} \\
\midrule
1 & \xmark & \xmark & \xmark & 84.72 \\
2 & \checkmark & \xmark & \xmark & 84.81\textsuperscript{\textnormal{↑0.09}} \\
3 & \xmark & \checkmark & \xmark & 85.58\textsuperscript{\textnormal{↑0.86}} \\
4 & \xmark & \xmark & \checkmark & 85.36\textsuperscript{\textnormal{↑0.64}} \\
5 & \checkmark & \checkmark & \xmark & 85.74\textsuperscript{\textnormal{↑1.02}} \\
6 & \checkmark & \xmark & \checkmark & 85.50\textsuperscript{\textnormal{↑0.78}} \\
7 & \xmark & \checkmark & \checkmark & 85.70\textsuperscript{\textnormal{↑0.98}} \\
8 & \checkmark & \checkmark & \checkmark & \textbf{86.04\textsuperscript{\textnormal{↑1.32}}} \\
\bottomrule
\end{tabular}
}
\end{table*}

\subsubsection{Effectiveness of various dimensions in AC-FA}
The results of the ablation studies between dimensions in AC-FA are shown in \cref{Ablation3}. AC-FA is composed of 3 dimensional modules: ACDA, ATDA and ASDA, aggregating features in channel, temporal, and spatial dimensions. Under the condition that SSTE module and F-RCL are adopted (No.1), one or two of the three dimensions are added randomly each time. According to No.2 to No.4, each dimension is conducive to the aggregation of features. Among them, spatial aggregation improves the lowest, since the clarity and stability of spatial relationships in skeleton data have enabled its features to be captured to a large extent passed through the SSTE module. Meanwhile, channel and temporal features have a relatively larger enhance, demonstrating the effectiveness of the dimensions in capturing dynamic dependencies as well as potential or indirect topology features. Besides, for pairwise combination of dimensions, channel and temporal, together with temporal and spatial presents significant synergistic effect, validating that temporal aggregation serves as a bridge to enhance the interplay between dynamic features in channel and spatial dimensions, which effectively capturing complex motion patterns and spatial-temporal dependencies. In contrast, channel and spatial combination are weakly synergistic due to the redundancy brought about by their jointly processing of static features. Moreover, the utilization of all three dimensions improves the accuracy by 0.54\%.

\begin{table*}[!t]
\centering
\caption{Ablation results of three dimensions aggregations in AC-FA on NTU-RGB+D 120 dataset's X-sub benchmark, using only joint modality as input.}
\label{Ablation3}
\scalebox{0.9}{
\begin{tabular}{ccccc}
\toprule
{\textbf{No.}} & \textbf{ACDA} & \textbf{ATDA} & \textbf{ASDA} & \textbf{Acc (\%)} \\
\midrule
1 & \xmark & \xmark & \xmark & 85.50 \\
2 & \checkmark & \xmark & \xmark & 85.68\textsuperscript{↑0.18} \\
3 & \xmark & \checkmark & \xmark & 85.70\textsuperscript{↑0.20} \\
4 & \xmark & \xmark & \checkmark & 85.60\textsuperscript{↑0.10} \\
5 & \checkmark & \checkmark & \xmark & 85.92\textsuperscript{↑0.42} \\
6 & \checkmark & \xmark & \checkmark & 85.75\textsuperscript{↑0.25} \\
7 & \xmark & \checkmark & \checkmark & 85.84\textsuperscript{↑0.34} \\
8 & \checkmark & \checkmark & \checkmark & \textbf{86.04\textsuperscript{↑0.54}} \\
\bottomrule
\end{tabular}
}
\end{table*}

\begin{table*}[!t]
\centering
\caption{Accuracy of our module combined with various GCN-based backbones. Only joint modality of NTU-RGB+D 120 is utilized as input.}
\label{CombinedAcc}
\scalebox{0.8}{
\begin{tabular}{ccccccc}
\toprule
\textbf{Method}        & \textbf{Params.} & \textbf{X-Sub (\%)} & \textbf{X-Set (\%)} \\ \midrule
ST-GCN \cite{yan2018spatial}       & 2.11M           & 83.62               & 85.32               \\
+ SF Head             & 2.12M           & 84.74\textsuperscript{\textbf{$\uparrow$1.12}} & 86.82\textsuperscript{\textbf{$\uparrow$1.50}} \\
2s-AGCN \cite{shi2019two}      & 3.80M           & 84.51               & 86.12               \\
+ SF Head             & 3.81M           & 85.56\textsuperscript{\textbf{$\uparrow$1.05}} & 87.00\textsuperscript{\textbf{$\uparrow$0.88}} \\
CTR-GCN \cite{chen2021channel}       & 1.46M           & 84.72               & 86.83               \\
+ SF Head             & 1.47M           & 86.04\textsuperscript{\textbf{$\uparrow$1.32}} & 87.76\textsuperscript{\textbf{$\uparrow$0.93}} \\
Info-GCN \cite{chi2022infogcn}      & 1.62M           & 85.24              & 86.45               \\
+ SF Head             & 1.63M           & 85.92\textsuperscript{\textbf{$\uparrow$0.68}} & 87.35\textsuperscript{\textbf{$\uparrow$0.90}} \\
HD-GCN \cite{Lee_2023_ICCV}       & 1.68M           & 85.42               & 87.36               \\
+ SF Head             & 1.69M           & 85.94\textsuperscript{\textbf{$\uparrow$0.52}} & 88.02\textsuperscript{\textbf{$\uparrow$0.66}} \\
Block-GCN \cite{zhou2024blockgcn}      & 1.30M           & 86.88              & 88.16               \\
+ SF Head             & 1.31M           & 87.25\textsuperscript{\textbf{$\uparrow$0.37}} & 88.64\textsuperscript{\textbf{$\uparrow$0.48}} \\
 ProtoGCN \cite{liu2024revealing}      & 4.34M           & 85.52              & 88.35               \\
 + SF Head             & 4.35M           & 85.95\textsuperscript{\textbf{$\uparrow$0.43}} & 88.68\textsuperscript{\textbf{$\uparrow$0.33}} \\
\bottomrule
\end{tabular}
}
\end{table*}

\subsection{Combination Effects with Other Backbones}
\label{sec4.4}
Our proposed approach is a highly encapsulated component that is fusible with a majority of GCN-based backbones. Its generality is evaluated by integrating it into seven mainstream backbone models on NTU RGB+D 120's x-sub and X-set benchmark. For fair comparison, only the joint modality is utilized as input to avoid the effect of multi-stream ensemble. The accuracies and parameter counts of the various methods are summarized in \cref{CombinedAcc}. As shown, integrating the SF-Head module consistently leads to a noticeable improvement in accuracy in all backbones, with an average increase of approximately 0.8\%. Notably, models with lower base backbone accuracy tend to benefit more significantly from SF-Head compared to those with higher initial accuracy. This effect may be attributed to the SF-Head’s ability to effectively enhance feature representation by focusing on subtle distinctions in ambiguous actions and mitigating information redundancy. By leveraging synchronized feature decoupling and adaptive refinement, SF-Head emphasizes critical local details and addresses challenges associated with misclassified samples. Additionally, SF-Head introduces only about 0.01M extra parameters, making the increase in computational cost negligible. This allows SF-Head to be integrated into large-scale systems without sacrificing efficiency.

It should be noticed that CTR-GCN is selected as our method's backbone in order to demonstrate the effectiveness of SF-Head in bridging the gap in feature representations between traditional methods and state-of-the-art (SOTA) approaches. Specifically, by integrating with CTR-GCN, our module (SF-Head) effectively enhances feature representation alignment and refinement, ensuring that both local and global information are well captured. In fact, using a backbone with an original accuracy greater than CTR-GCN would naturally yield higher performance. Yet our goal is to show that SF-Head enhances existing architectures by addressing the gaps in feature representation, rather than simply relying on the inherent strengths of a SOTA backbone.

\subsection{Performance and Visualization of Ambiguous Actions}
\subsubsection{Accuracy enhancement on Ambiguous Actions}
This segment verifies the strong recognition ability of SF-Head in ambiguous actions. 

\begin{figure*}[!t]
    \centering
    \includegraphics[scale=0.45]{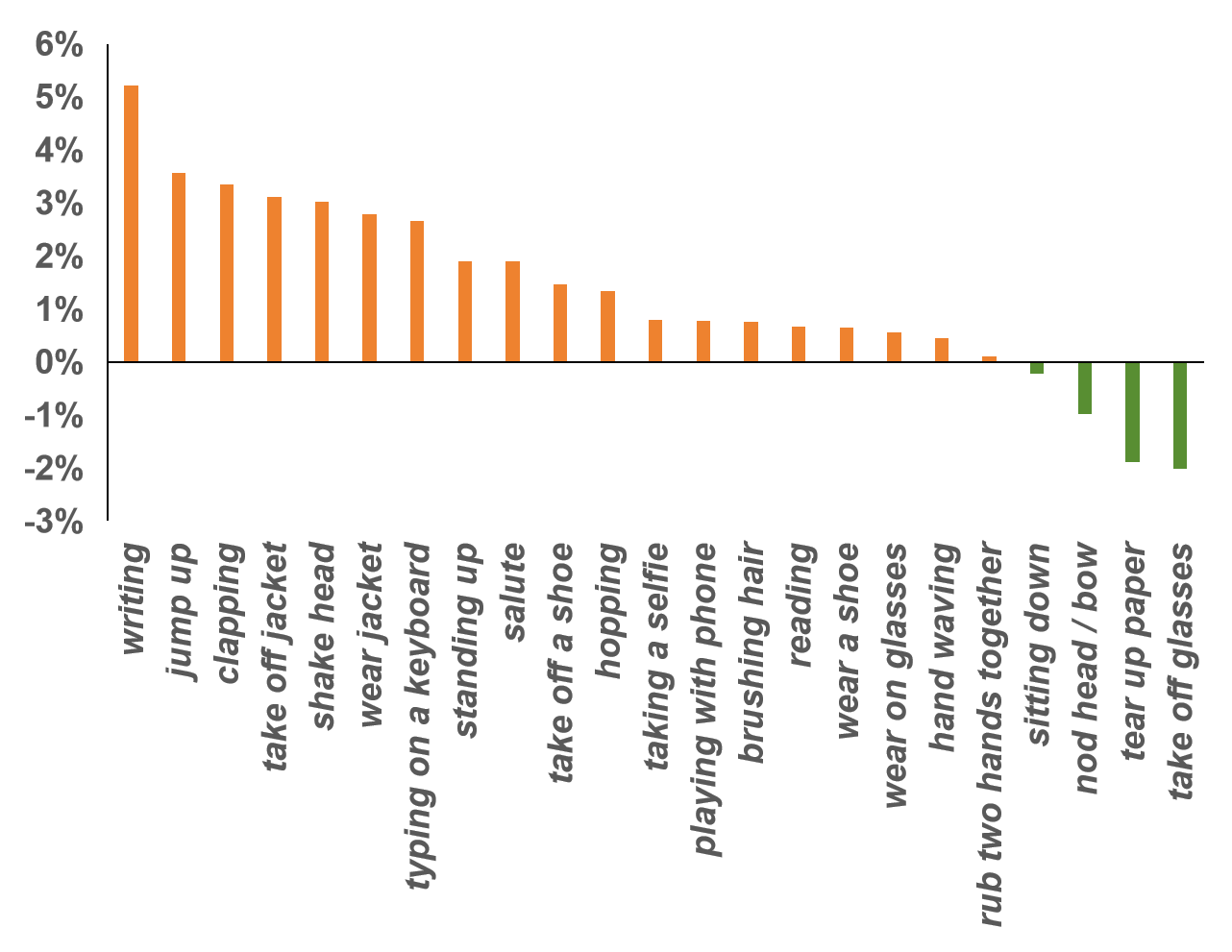}
    \caption{Accuracy of 23 ambiguous actions (9 groups) in descending order. Group 1: writing, typing on a keyboard, playing with phone, reading; Group 2: jump up, hopping; Group 3: clapping, rub two hands together, tear up paper; Group 4: take off jacket, wear jacket; Group 5: shake head, nod head / bow; Group 6: standing up, sitting down; Group 7: salute, taking a selfie, brushing hair, hand waving; Group 8: take off a shoe, wear a shoe; Group 9: wear on glasses, take off glasses}
    \label{23classes}
\end{figure*}

On NTU-RGB+D 120's X-sub benchmark, 9 groups of ambiguous actions (23 actions in total) are selected. By subtracting the accuracy of each selected action in our model with the CTR-GCN backbone\cite{chen2021channel}, the differences are displayed in \cref{23classes}, arranging in descending order from left to right. Among them, the most typical group is "writing, typing on a keyboard, playing with phone, reading", which are enhanced by 5.21\%, 2.65\%, 0.78\%, 0.67\% respectively. Another typical group is "take off jacket, wear jacket", which are enhanced by 3.12\% and 2.78\% respectively. Notably, the accuracy increments of most similar action groups are significantly higher than the enhancement of 1.7\% (from 88.9\% to 90.6\%, presented in SOTA) in global accuracy. This indicates that SF-Head stands out in identifying localized features among ambiguous action groups.

\begin{figure*}[ht!]
    \centering
    \includegraphics[scale=0.35]{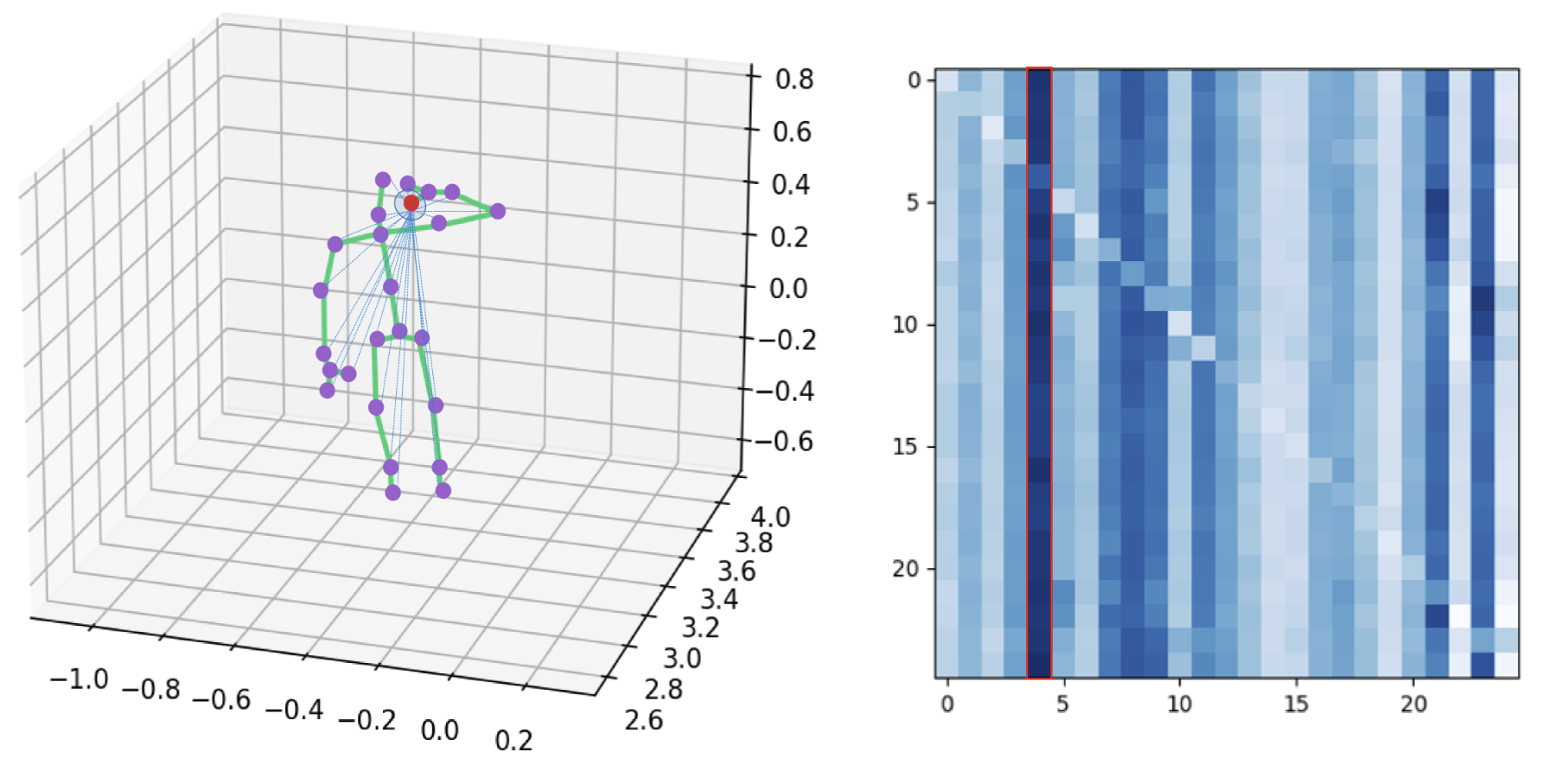}
    \caption{Adaptive learning matrix of the action "salute" on NTU-RGB+D 60 dataset}
    \label{human vis}
\end{figure*}

\begin{figure*}[ht!]
    \centering
    \includegraphics[scale=0.35]{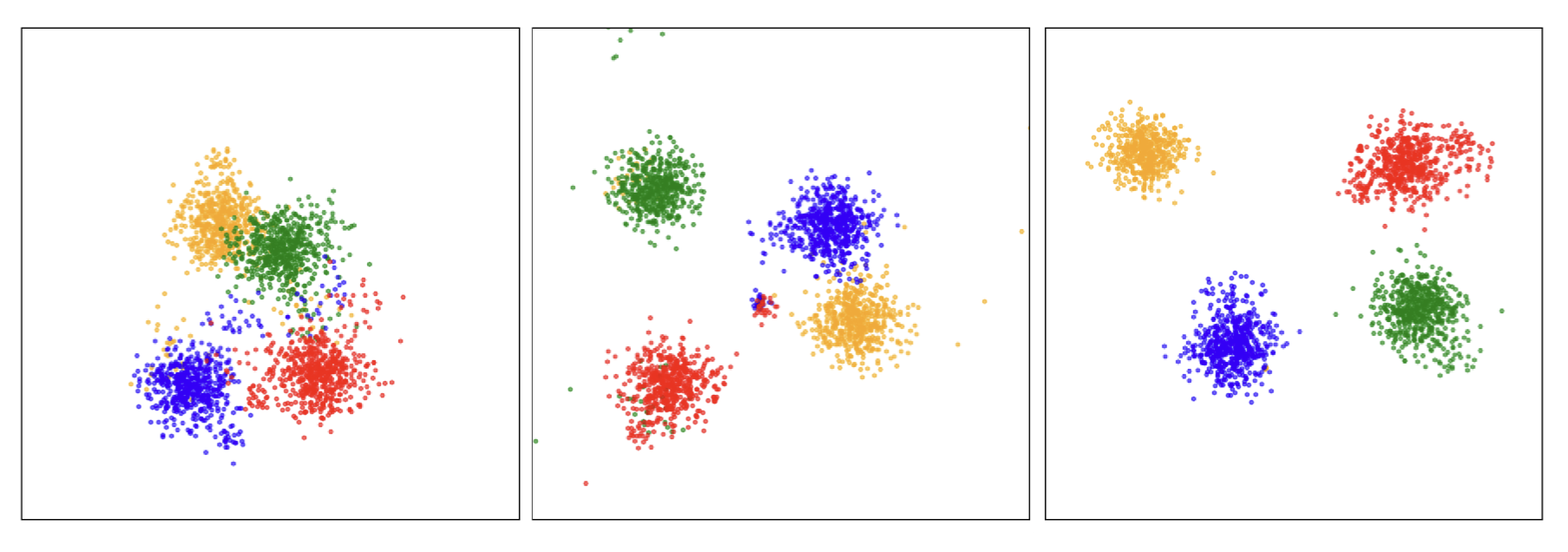}
    \caption{Representations of long-term action sequences in the NTU-RGB + D 120 Xset dataset using only the backbone (left), backbone + our module w.o. F-RCL (middle), and backbone + our module (right). Each color represents one unique feature.}
    \label{clusters}
\end{figure*}

\subsubsection{Validation of Potential Topological Structures}
In order to validate that SF-Head module is capable of aggregating joint dependencies effectively, \cref{human vis} is displayed to show the skeleton structure and the adaptive learning process of the adjacency matrices concerning the action "salute" on NTU-RGB+D 120 dataset. The lighter the color, the weaker the connection between two joints, suggesting greater feature interactions and dependencies. Notably, the 4th row of the adjacency matrix, corresponding to the red point in the skeleton, is significantly deeper in color compared with other rows. This shows that the highlighted red joint is the most crucial part in distinguishing "salute" from other actions, which adapts to semantics in reality because connecting the joint of the hand to the head serves as the most essential feature for recognizing salute movements. Therefore, the effectiveness of our model in refined feature extraction and potential topological structures learning has been demonstrated.

\subsubsection{Visualization of Features in 2D Plane}
\label{sec4.5.3}
Four long-term action categories, "Wear on glasses", "Take off glasses", "Drop", and "Pick up", are selected to visualize their distribution in the feature space. T-distributed Stochastic Neighbour Embedding (t-SNE) method \cite{van2008visualizing} is utilized to map the features in a 2D plane. As shown in \cref{clusters}, the spacing between point clusters gradually increases from left to right, indicating a reduction in intra-class overlap and an enhancement in inter-class separability. The base backbone (left) struggles to differentiate between ambiguous actions, leading to overlapping clusters. By introducing our module without F-RCL (middle), noticeable improvement in cluster separability is observed, while further applying the F-RCL (right) helps to effectively eliminate feature noise and enhances both feature consistency and diversity. Therefore, SF-Head outperforms the base backbone and SF-Head w.o. F-RCL in distinguishing their similarities, validating the contributions of SSTE, AC-FA and F-RCL in our proposed module.

\afterpage{
\begin{table*}[!htbp]
\centering
\caption{Comparison of top-l accuracy (\%) on NTU-RGB+D 60, NTU-RGB+D 120 and NW-UCLA. The best one is in bold and the second one is underlined.}
\label{SoTA}
\scalebox{0.7}{
\begin{tabular}{llcccccc}
\toprule[1.5pt]
\textbf{Method}          & \textbf{Publication} & \multicolumn{2}{c}{\textbf{NTU RGB+D}} & \multicolumn{2}{c}{\textbf{NTU RGB+D 120}} & \textbf{NW-UCLA} \\ 
\cmidrule(lr){3-4} \cmidrule(lr){5-6}
                         &                      & \textbf{X-Sub} & \textbf{X-View}    & \textbf{X-Sub}    & \textbf{X-Set}    &                   \\ 
\midrule
ST-GCN \cite{yan2018spatial}      & AAAI'2018             & 81.5          & 88.3              & 70.7                & 73.2                & -                 \\ 
AS-GCN \cite{li2019actional}    & CVPR'2019             & 86.8          & 94.2              & 77.9                & 78.5                & -                 \\ 
2s-AGCN \cite{shi2019two}    & CVPR'2019             & 88.5          & 95.1              & 82.9             & 84.9             & -                 \\ 
AGC-LSTM \cite{si2019attention}  & CVPR2019             & 89.2          & 95.0              & -                & -                & 93.3              \\ 
Dynamic GCN \cite{ye2020dynamic} & ACMMM'2020      & 91.5          & 96.0              & 87.3             & 88.6             & -                 \\ 
SGN \cite{zhang2020semantics}           & CVPR'2020             & 89.0          & 94.5              & 79.2             & 81.5             & -                 \\ 
Shift-GCN \cite{cheng2020skeleton} & CVPR'2020            & 89.7          & 96.0              & 85.9             & 87.6                & 94.6              \\ 
MS-G3D \cite{liu2020disentangling}      & CVPR'2020             & 91.5          & 96.2              & 86.9             & 88.4             & -                 \\ 
DDGCN \cite{korban2020ddgcn}       & ECCV'2020             & 91.1          & 97.1     & -                & -                & -                 \\ 
DC-GCN+ADG \cite{cheng2020decoupling}  & ECCV'2020             & 90.8          & 96.6              & 86.5             & 88.1             & 95.3              \\ 
MST-GCN \cite{chen2021multi}    & AAAI'2021             & 89.0          & 95.1              & 82.8             & 84.5             & 87.5              \\ 
Skeletal-GNN \cite{zeng2021learning} & ICCV'2021      & 91.6          & 96.7              & 87.5             & 89.2             & -                 \\ 
CTR-GCN \cite{chen2021channel}    & ICCV'2021             & 92.4          & 96.8              & 88.9             & 90.6             & 96.5              \\
% Js-CTR-GCN \cite{mstgcn}    & ICCV'2021             & 90.1          & 94.6              & 84.9             & 87.0             & -              \\
ShiftGCN++ \cite{cheng2021extremely}    & ICCV'2021             & 90.5          & 96.3              & 85.6             & 87.2             & -              \\ 
STF \cite{ke2022towards}           & AAAI'2022             & 92.5          & 96.9              & 88.9             & 89.9             & -                 \\ 
Ta-CNN \cite{xu2022topology}      & AAAI'2022             & 90.4          & 94.8              & 85.4             & 86.8             & 96.1              \\ 
Info-GCN(4-ensemble) \cite{chi2022infogcn}           & CVPR'2022             & 92.7          & 96.9             & 89.4             & 90.7             &  96.6                 \\
Info-GCN(6-ensemble) \cite{chi2022infogcn}           & CVPR'2022             & 93.0          & 97.1              & 89.8             & 91.2             &  97.0                 \\
EfficientGCN-B4 \cite{song2022constructing} & TPAMI'2022             & 91.7          & 95.7              & 88.3             & 89.1             & -                 \\ 
HD-GCN(4-ensemble) \cite{Lee_2023_ICCV}           & CVPR'2023             &  93.0          & 97.0              & 89.8           &  91.2             & 96.9                 \\
HD-GCN(6-ensemble) \cite{Lee_2023_ICCV}           & CVPR'2023             &  93.4        & 97.2              & 90.1            &  91.6             & \underline{97.2}                 \\
HiCLR \cite{zhang2023hierarchical}           & AAAI'2023             & 90.4          & 95.7              & 85.6             & 87.5             & -                 \\
GAP \cite{xiang2023generative}           & ICCV'2023             & 92.9          &  97.0              &  89.9             & 91.1             & -                 \\
RVTCLR+ \cite{zhu2023modeling}           & ICCV'2023             & 87.5          & 93.9              & 82.0             & 83.4             & -                 \\
3s-HYSP \cite{zhu2023modeling}           & ICLR'2023             & 89.1          & 95.2              & 84.5             & 86.3             & -                 \\
 DS-GCN \cite{xie2024dynamic}           & AAAI'2024             & 93.1          & \underline{97.5}          & 89.2          & 91.1             & -                 \\
BlockGCN \cite{zhou2024blockgcn}           & CVPR'2024             & 93.1          & 97.0          & 90.3          & 91.5             & 96.9                 \\
 SkateFormer \cite{do2025skateformer}           & ECCV'2024             & \underline{93.5}          & \textbf{97.8}          & 89.8          & 91.4             & \textbf{98.3}                 \\
 ProtoGCN(4-ensemble) \cite{liu2024revealing}           & \multicolumn{1}{c}{-}             & \underline{93.5}          & \underline{97.5}          & 90.4          & 91.9             & -                 \\
 ProtoGCN(6-ensemble) \cite{liu2024revealing}           & \multicolumn{1}{c}{-}             & \textbf{93.8}          & \textbf{97.8}          & \underline{90.9}          & \textbf{92.2}             & -                 \\
\midrule
\textbf{Ours(4-ensemble)}            & \multicolumn{1}{c}{-}                    & 93.1              & 96.9             & 90.6             & 91.7             & 96.9\\ 
 \textbf{Ours(6-ensemble)}            & \multicolumn{1}{c}{-}                    & \underline{93.5}              & 97.1             & \textbf{91.0}             & \underline{92.0}             & \underline{97.2}\\ 
\bottomrule[1.5pt]
\end{tabular}
}
\end{table*}
}

\subsection{Comparison with the SOTA Methods}

Aiming at validating the effectiveness of our approach, we evaluate the model compared to state-of-the-art networks on four datasets: NTU-RGB+D 60, NTU-RGB+D 120, NW-UCLA and PKU-MMD I. Notably, most of the frameworks are employed in multi-stream fusion. Thus, to achieve fair comparisons, the CTR-GCN framework\cite{chen2021channel} is applied, with the six-stream ensemble strategy \cite{chi2022infogcn}.

\subsubsection{NTU-60/120 and NW-UCLA}
According to the results reported in \cref{SoTA}, it can be observed that on NTU-RGB+D 60 X-Sub and on both X-Sub and X-Set of NTU-RGB+D 120, our proposed model surpasses all current published methods, which reaches accuracies of 93.5\%, 91.0\%, and 92.0\% respectively. On NW-UCLA, our model is second to SkateFormer\cite{do2025skateformer}, being consistent with HD-GCN\cite{Lee_2023_ICCV}, also showing excellent performance.

In addition, it is worth noting that we rather focus more on our model's ability to distinguish similar actions than accuracy, which results in more outstanding performance on complicated datasets (e.g. NTU-RGB+D 120) and stronger applicability in reality.

\subsubsection{PKU-MMD I}
In \cref{SOTA2}, we compare our model to other SOTA methods on the PKU-MMD I
dataset. Our model is second to 3s-HYSP\cite{zhu2023modeling}, also showing excellent recognition performance.

\afterpage{
\begin{table*}[hb!]
\centering
\caption{Classification accuracy(\%) comparisons against SOTA methods on PKU-MMD I. The best one is in bold and the second one is underlined.}
\label{SOTA2}
\scalebox{0.8}{
\begin{tabular}{ccc}
\toprule
\textbf{Method}        & \textbf{Publication} & \textbf{Acc (\%)} \\ \midrule
 ST-GCN\cite{yan2018spatial}              & AAAI'2018 &  84.1   \\
   $\text{MS}^2L$ \cite{lin2020ms2l} & ACM MM'2020 & 85.2  \\
   3s-HYSP\cite{zhu2023modeling} &  ICLR'2023  & \textbf{96.2} \\
 Ours  & \multicolumn{1}{c}{-}  & \underline{95.9} \\

\bottomrule
\end{tabular}
}
\end{table*}
}

\section{Conclusion}

In this paper, we present the Synchronized and Fine-grained Head (SF-Head), an innovative lightweight, plug-and-play module designed to enhance the discriminative power of skeleton-based ambiguous action recognition models. SF-Head employs the SFFE module to synchronized spatial-temporal feature extraction with F-RL to balance these two types of features, followed by the AC-FA module to focus on local details without sacrificing global context with F-CL to preserve the alignment between the aggregated and original features, which is crucial for distinguishing visually similar actions.

Extensive experimental evaluations were conducted on four widely-used benchmark datasets including NTU RGB+D 60, NTU RGB+D 120, NW-UCLA and PKU-MMD I. The results demonstrate that our proposed method significantly outperforms existing approaches, especially in ambiguous actions. Notably, SF-Head adds fewer than 0.01M parameter and incurs no additional cost during inference, which ensures its practicality for real-world applications. Future work will focus on further optimizing SF-Head for more efficient feature refinement processes and exploring its integration with other modalities, such as RGB imagery, to fully exploit multi-modal data for action recognition tasks.

\subsection*{AUTHOR CONTRIBUTIONS}
\textbf{Hao Huang:} Data curation; methodology; validation; visualization; writing - original draft; writing - review \& editing. \textbf{Yujie Lin:} Data curation; validation; writing - original draft; writing - review \& editing. \textbf{Siyu Chen:} investigation; visualization; writing - original draft; writing - review \& editing. \textbf{Haiyang Liu:} formal analysis; funding acquisition; project administration; supervision.

\subsection*{ACKNOWLEDGEMENTS}
This work was supported by the Fundamental Research Funds for the Central Universities No.2023JBZY035.

\subsection*{CONFLICT OF INTEREST STATEMENT}
The authors declare no conflicts of interest.

\subsection*{DATA AVAILABILITY STATEMENT}
The data that support the findings of this study are available from the corresponding author upon reasonable request.

\subsection*{ORCID}
\textit{Hao Huang} \orcidlink{0009-0001-9791-4452} \href{https://orcid.org/0009-0001-9791-4452}{https://orcid.org/0009-0007-4813-8112}

\bibliographystyle{splncs04}
\bibliography{main}
\end{document}